\documentclass{article}

\PassOptionsToPackage{numbers,compress}{natbib}
\usepackage[preprint]{neurips_2026}

\usepackage[utf8]{inputenc} % allow utf-8 input
\usepackage[T1]{fontenc}    % use 8-bit T1 fonts
\usepackage{hyperref}       % hyperlinks
\usepackage{url}            % simple URL typesetting
\usepackage{booktabs}       % professional-quality tables
\usepackage{amsfonts}       % blackboard math symbols
\usepackage{nicefrac}       % compact symbols for 1/2, etc.
\usepackage{microtype}      % microtypography
\usepackage{xcolor}         % colors
\usepackage{float}
\usepackage[skins]{tcolorbox}
\usepackage{caption}
\usepackage{fancyvrb}
\usepackage{subcaption}

\usepackage{listings}

\usepackage{graphicx}
\usepackage{listings}
\usepackage{multirow}
\usepackage{placeins}
\usepackage{tikz}
\usepackage{varwidth}

\title{The Stanford EDGAR Filings Dataset: Reconstructing U.S. Corporate and Financial Disclosures into Layout-Faithful and Token-Efficient Pretraining Data}

\author{%
  Nick Bettencourt \\
  Department of Applied Statistics and Data Science\\
  University of California, Los Angeles\\
  \texttt{nbetts@g.ucla.edu} \\
  \And
  Xiaowei Ding* \\
  School of Information Management\\
  Nanjing University\\
  \texttt{dingxiaowei@nju.edu.cn} \\
  \And
  Kay Giesecke* \\
  Advanced Financial Technologies Lab\\
  Stanford University\\
  \texttt{giesecke@stanford.edu} \\
}

\begin{document}

\maketitle

\begingroup
\renewcommand{\thefootnote}{}
\footnotetext{\hspace*{-1.8\parindent}*\hspace{0.05em}Corresponding Authors\newline\textbf{AI usage disclosure.}
AI systems were used as methodological tools and evaluated systems in this work: Mistral OCR 3 was used for PDF-derived filing extraction; GPT-5.4/GPT-5.5 were used in benchmark construction and table-reconstruction workflows; and LLM/OCR systems were evaluated in EDGAR-OCR, EDGAR-Forecast, and the HTML table reconstruction benchmark. The human authors reviewed AI-assisted artifacts, citations, figures, model outputs, and scientific claims, and take full responsibility for the paper. LLMs were also used for routine drafting, editing, formatting, and code assistance.}
\endgroup

\begin{abstract}
  \noindent As high-quality public web corpora become increasingly exhausted, clean long-context documents have become a scarce and expensive source of training data for large language models (LLMs). Existing long-context corpora are often proprietary and costly to acquire, synthetically generated, or concentrated in narrow domains such as programming. We introduce the Stanford EDGAR Filings Dataset (SEFD), an open reconstruction of SEC filings into layout-faithful MultiMarkdown for financial language modeling and evaluation. SEFD makes audited financial statements, risk disclosures, ownership reports, accounting notes, and market-moving event filings usable as long-context pretraining data and as a basis for financial reasoning, forecasting, compliance, and document understanding. The resulting corpus is token-efficient, model-ready, and has less than 0.1\% overlap with Common Crawl-derived corpora. We release SEFD-v1, a 152B-token initial public snapshot, and provide corpus-level analyses of a larger 18.5M-filing archive estimated at 550B tokens. We further introduce two SEFD-derived benchmarks: EDGAR-Forecast, which evaluates filing-grounded numerical forecasting after model knowledge cutoffs, and EDGAR-OCR, which evaluates transcription of complex financial tables.
\end{abstract}

\section{Introduction}

The "scaling era" of modern LLMs has been defined by 'bigger is better' -- yielding the belief that more compute, more tokens, and more parameters will produce better capabilities. Yet, as evidenced by recent models such as GPT-4.5 and Llama 4 Behemoth \cite{openai:gpt45,meta:llama4}, this brute-force recipe is producing diminishing returns even as pretraining data scales upwards of 30 trillion tokens \cite{meta:llama4}. In response, "quality over quantity" has emerged as a compelling alternative. Microsoft’s Phi line explicitly attributes its strength to 'textbook-quality' data curation, demonstrating competitive performance against larger models on common benchmarks \cite{gunasekar2023textbooks}. Even within standard pretraining, results indicate that improved data and token allocation can allow smaller models to outperform larger ones. As model architectures and training recipes increasingly converge across the industry, the data itself is becoming the primary differentiator of performance, underscoring that targeted, meticulous data curation yields the most efficient reduction in validation loss per FLOP.

\noindent EDGAR, the U.S. Securities and Exchange Commission's (SEC) public disclosure archive, is one of the largest public repositories of factual, business-oriented long-context data, processing roughly 4,700 filings per day and accommodating about 40,000 new filers per year \cite{sec:about-edgar}. Yet, it has remained largely untapped as complete pretraining data because its 18.5M filings present broad parsing challenges stemming from how disclosures are created, rendered, and filed. In this work, we introduce the \textbf{Stanford EDGAR Filings Dataset (SEFD)}, a highly curated corpus derived from the full EDGAR archive (1994--present) \cite{sec:access-edgar}. The full SEFD corpus is estimated at 550B tokens, and we release SEFD-v1 as a 152B-token initial public snapshot covering filings from January 2022 through June 2025. Unlike prior EDGAR-derived corpora, which often release cleaned plain-text extractions while dropping layout-heavy elements such as tables and indentation structures \cite{wang2024beancounter, loukas2021edgarcorpus}, SEFD reverse-engineers these visual layouts into a layout-faithful and token-efficient representation for LLM pretraining. These cues carry financial meaning. Indentation disambiguates statement hierarchies, merged headers connect periods and segments without repeated text, and numeric-cell reconstruction reattaches signs, currency symbols, and percent symbols to the values they modify. Flattening EDGAR tables can create ambiguity by detaching values from labels, duplicating headers, or flipping accounting signs. By preserving this structure at scale, SEFD enables long-context pretraining and filing-grounded applications such as forecasting, compliance review, accounting question answering, table extraction, document OCR evaluation, and agentic financial analysis. Human evaluation shows our rule-based methodology achieves high structural and semantic accuracy ($>$99\%)\footnote{Translates to roughly 1 understanding-breaking error per 100 pages}, and SEFD has less than 0.1\% overlap with Common Crawl-derived corpora such as C4 \cite{commoncrawl, raffel2020t5,wang2024beancounter}.

\noindent Finally, to evaluate contamination-resistant downstream reasoning and extraction, we introduce two benchmarks: \textbf{EDGAR-Forecast} and \textbf{EDGAR-OCR}. EDGAR-Forecast tests agentic financial reasoning in a sandbox with no internet access, where models receive only the preceding 5 years of a company's SEC filings and forecast numerical values in 2026 Form 10-Q filings released after the evaluated models' knowledge cutoffs. The strongest model we tested, GPT-5.5 \cite{openai:gpt55}, achieves 51.8\% on 250 questions across 50 companies. EDGAR-OCR measures small, high-throughput LLMs at transcribing complex financial tables into HTML. Its 300 tables are synthetically transformed from real SEC filings by replacing entities, dates, labels, and values while preserving layout and verified arithmetic relationships, reducing memorization and retrieval contamination. Qwen3.6-35B-A3B \cite{qwen:qwen36} achieves the top score at 75.78\%. Together, these benchmarks show that SEFD's high-fidelity parsing expands EDGAR beyond retrieval-augmented search, supporting model evaluation for financial, business, and accounting tasks as well as pretraining and RLVR-style dataset construction.

\section{Related Work}

\paragraph{Large-scale pretraining corpora.}
Most large open pretraining corpora are assembled from heterogeneous web sources such as Common Crawl, including variants such as C4, RefinedWeb, and FineWeb \cite{commoncrawl,raffel2020t5,penedo2023refinedweb,penedo2024fineweb}. These datasets provide scale, but require aggressive filtering to remove spam, boilerplate, duplication, and incoherent text. They also typically flatten document structure, discarding layout cues such as tables, indentation, and visual hierarchy. Although the SEC EDGAR Database is entirely public, its filings pose a different kind of construction challenge -- valuable financial content is embedded in noisy HTML, legacy formatting conventions, and layout-heavy structures that are poorly suited to be captured by standard text extraction pipelines.

\paragraph{Financial corpora and EDGAR tooling.}
Financial LLM corpora such as BloombergGPT's FinPile and BeanCounter \cite{bloomberggpt,wang2024beancounter} incorporate SEC filings as an important source of domain text. FinPile combines filings with news, transcripts, press releases, and proprietary financial content, but is closed and only partially EDGAR-derived. BeanCounter is the closest large-scale open comparison, releasing a 159B-token corpus from high-volume EDGAR forms such as 10-K, 10-Q, and 8-K, while removing numeric tables rather than preserving them. Separately, open-source tools such as EdgarTools \cite{edgartools} provide useful access to metadata, XBRL facts, and filing sections for retrieval-oriented applications. These resources make financial filings more usable as text or retrieval targets, but they are not designed to produce corpora for pretraining or layout-faithful evaluation. SEFD instead targets a layout-faithful MultiMarkdown \cite{multimarkdown} representation that preserves tables, indentation, and visual relationships important for financial reasoning.

\paragraph{Document and financial agent benchmarks.}
OmniDocBench \cite{omnidocbench} provides a broad benchmark for document parsing and OCR capabilities, including layout, table, formula, and reading-order recovery, and is the closest methodological inspiration for EDGAR-OCR. Vals AI's FinanceAgent v1.1 \cite{vals:financeagent} evaluates agentic financial analysis over realistic analyst tasks involving SEC filings, retrieval, quantitative reasoning, and tool use; its public harness relies on EDGAR search and plain HTML-to-text parsing rather than layout-faithful filing reconstruction. FinanceAgent motivates EDGAR-Forecast, but EDGAR-Forecast focuses specifically on predicting future financial outcomes from historical SEC filings, rather than evaluating broad analyst workflows over already-available filings and public data.

\section{Dataset Construction}

\noindent The structural mess of raw SEC filings is best understood as the byproduct of a regulatory workflow optimized for compliant visual disclosure rather than machine-readable text. Clean reconstruction can often yield a $>$99\% reduction in token count relative to the raw source, because much of modern filing HTML consists of presentation scaffolding rather than filer-authored content, a trend that has intensified as disclosure production has moved into enterprise platforms and filing agents. Gaining traction in the mid-2000s, filings are now typically produced through enterprise disclosure-management platforms and filing agents such as Workiva, Donnelley Financial Solutions (DFIN), Toppan Merrill, and others.\footnote{Public filing-agent volume aggregations list Donnelley Financial Solutions, Toppan Merrill, and EdgarAgents among the highest-volume filing agents in 2025, followed by Quality Edgar Solutions, WallStreetDocs, and Workiva.} These systems assemble financial statements, narrative disclosures, XBRL tags, and reporting templates from accounting and disclosure workflows, then render them into SEC-compliant documents. In order to maintain a stable, print-like appearance across browsers, EDGAR restricts filing HTML to accepted tags and HTML~3.2/4.0-era attributes, encouraging table-based layout (\texttt{<table>}, \texttt{<tr>}, \texttt{<td>}), presentational markup, and explicit whitespace (\texttt{\&nbsp;}, \texttt{<br>}) \cite{sec:efm-vol2, w3c:html32}. Browsers can visually glue these markup fragments together, but parsers must infer which separated pieces are semantically connected.

\begin{figure}[H]
    \centering
    \includegraphics[width=1\linewidth]{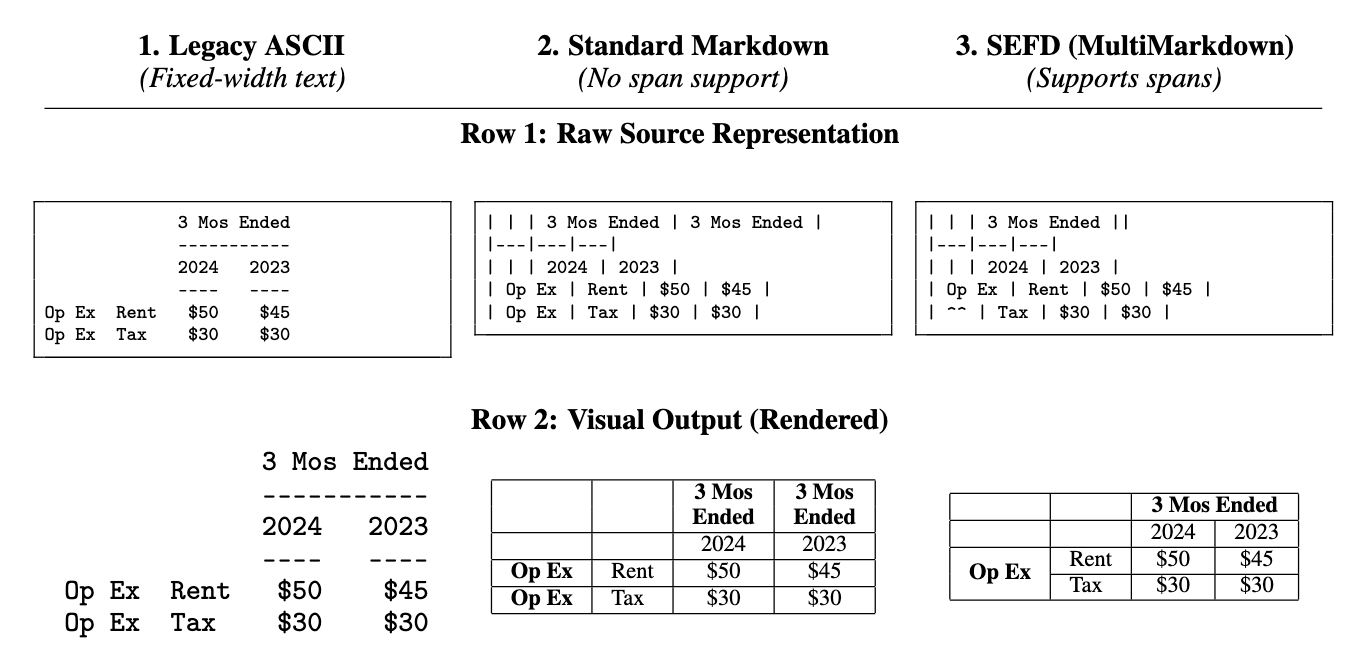}
    \caption{Comparison of table representations. \textbf{(1)} ASCII relies on whitespace for alignment, capturing most visual elements but is not token-efficient (though this depends mostly on the tokenizer's whitespace handling). \textbf{(2)} Standard Markdown, the most popular format for pretraining, fails to represent complex hierarchies, necessitating redundant data or empty cells to maintain alignment. \textbf{(3)} MultiMarkdown natively supports complex table structures via \texttt{||} (column spans) and \texttt{\char`^\char`^} (row spans), faithfully preserving the semantic grouping of financial data.}
    \label{fig:table_formats}
\end{figure}

To make those connections explicit, SEFD employs a 'visual-first' parsing methodology, reconstructing spatial semantics rather than relying on naive DOM-based text extraction. By treating the rendered document as a 2D coordinate grid instead of a linear tag hierarchy, the parser reattaches fragmented text spans and resolves visual indentation before serialization. Data ingress begins by parsing the SGML header for the \texttt{CONFORMED SUBMISSION TYPE}, routing filings across 33 specialized XML schemas (Forms 4, 13F, N-PORT, etc.) and over 350 filing types, including 10-K, 10-Q, and 8-K variants. SEFD uses MultiMarkdown (MMD) \cite{multimarkdown} because it represents complex table logic token-efficiently while remaining a small extension of Markdown, a widely used presentation language. MMD encodes horizontal spans with consecutive pipe delimiters (\texttt{||}) and vertical spans with caret markers (\texttt{\^{}\^{}}), preserving merged-cell structure without HTML tags. Figure \ref{fig:table_formats} contrasts legacy ASCII, standard Markdown, and MultiMarkdown table representations. Finally, SEFD applies deterministic normalization during construction: removing commas from numerical values, dropping non-semantic artifacts such as isolated page numbers, and prepending metadata such as CIK and SIC code from the \texttt{<SEC-HEADER>} (Listing~\ref{lst:sgml_header}). These post-processing steps apply universally, while initial reconstruction varies by presentation language, sometimes within the same filing.

\subsection{Plaintext}

\noindent While most of SEFD consists of HTML or XML, the archive contains a significant tail of legacy ASCII documents, predominantly from filings before the widespread adoption of HTML in the early 2000s\footnote{The gradual shift away from ASCII filings coincides with the SEC's acceptance of EDGAR filings in HTML starting on June 28, 1999}, as well as form types that remained plaintext-based longer. These include Form NSAR, which used a key-value ASCII pair system until 2018, and legacy Form 13F holdings reports filed before the 2013 XML mandate. Some plaintext filings also use legacy SGML tags such as \texttt{<TABLE>} and \texttt{<S>} as EDGAR-specific formatting controls for fixed-width tables. For all such documents, our parsing strategy shifts from reconstruction to preservation. Because these filings rely entirely on fixed-width characters and whitespace to denote tables and alignment, applying any normalization to the whitespace would destroy the document's structure. Our pipeline first identifies these files by a regex-based check for a lack of structural HTML tags (\texttt{<html>}, \texttt{<div>}, \texttt{<p>}), then moves to parsing headers and submission metadata. Once isolated, the text body is wrapped in a Markdown code fence block; the only normalization collapses 3 or more consecutive isolated line breaks into 2, preserving pagination gaps while improving token efficiency.\footnote{A small subset of 'Paper' filings contain a \texttt{<PAPER>} SGML tag rather than substantive disclosure text, usually due to physical-delivery requirements or hardship exemptions. SEFD parses the \texttt{<SEC-HEADER>} metadata and reformats the notice into a Markdown blockquote, preserving regulatory history when digital text is absent.}

\subsection{HTML}

As indicated in Figure~\ref{fig:dataset-source-format-comparison}, HTML-derived tokens account for roughly 62\% of the SEFD corpus. HTML originated as an SGML application with Document Type Definitions (DTDs), but the browser wars of the 1990s incentivized a permissive parsing model known informally as 'Tag Soup', wherein the DOM reflects a repaired \textit{interpretation} rather than a strictly valid structure. In EDGAR, the problem is exacerbated by regulatory constraints that limit filers to a narrow legacy subset of HTML and forbid most external dependencies and external CSS files, removing many of the layout tools normally used to typeset financial tables. Filing software therefore relies on 'layout engineering,' mimicking modern CSS through limited HTML syntax and decoupling document structure from content. SEFD addresses this by treating HTML parsing as a reconstruction of the rendered coordinate system rather than a traversal of the syntactic tree. Beyond table geometry, SEFD also preserves document-level visual signals. CSS margins, padding, fixed-width empty columns, and non-breaking spaces are converted into discrete indentation levels, while inline styling is preserved by inserting paired placeholder tokens around tags before DOM parsing. Appendix~\ref{app:html-layout-signals} provides additional details.

\paragraph{Tables}\label{sec:tables}

The most pervasive artifact of 'layout engineering' is the fragmentation of quantitative values. In financial reporting, numbers are typically right-aligned, and decimals are aligned to a common x-coordinate within the column. However, legacy HTML has no native “decimal tab” for aligning on a character. A right-alignment breaks accounting formats because parentheses, which are used to represent negative numbers, introduce extra glyphs that shift the digits and disrupt column alignment. Solving this, filers universally adopt a "Three-Column Hack." Single numerical values are exploded across three distinct table cells:

\vspace{-0.2em}

\begin{enumerate}

\item \textbf{Prefix} or \textbf{Gutter} Column: Often defined with \texttt{width="1\%"}, holding floating currency symbols (\$) or opening parentheses.
\item \textbf{Value} Column: Containing the integer and decimal.
\item \textbf{Suffix} Column: Holding closing parentheses or percentage signs.

\end{enumerate}

\begin{figure}
    \centering
    \includegraphics[width=1\linewidth]{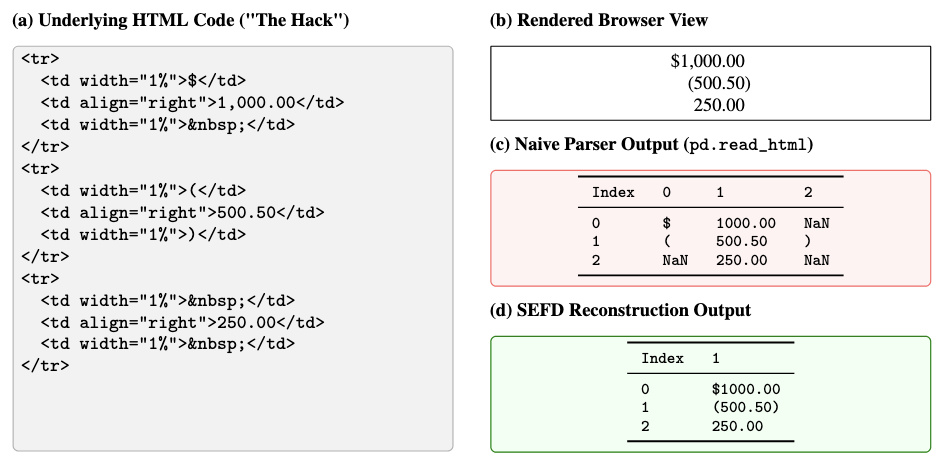}
    \caption{\textbf{The "Three-Column Hack."} EDGAR tables split displayed numbers across prefix, value, and suffix cells for decimal alignment; standard parsers separate these parts, while SEFD reconstructs them.}
    \vspace{-1em}
    \label{fig:three_column_hack}
\end{figure}

\noindent \textbf{Symbol Reconstruction.} Understanding this creation process, SEFD employs a bidirectional rule-based merging algorithm. The parser scans the table grid for ``modifier columns,'' consisting exclusively of currency symbols, percentages, or parenthetical fragments (e.g., \texttt{)}, \texttt{)\%}, \texttt{)bp}), and collapses them into adjacent value columns. Prefixes (e.g., ``\$'') are merged rightward and suffixes (e.g., ``\%'') leftward, using lookahead and lookbehind checks to ensure the target column contains compatible numeric data. Because footnote markers and decorative spacing can superficially resemble these modifiers, the classification logic is context-aware: it distinguishes semantic modifiers (e.g., a leading ``\$'' indicating currency) from structural artifacts (e.g., an orphaned currency-symbol column with no filled value to attach). This prevents false positives and lets reconstruction target quantitative polarity rather than accidental adjacency.

\vspace{0.2em}

\begin{figure}[H]
    \centering
    \includegraphics[width=1\linewidth]{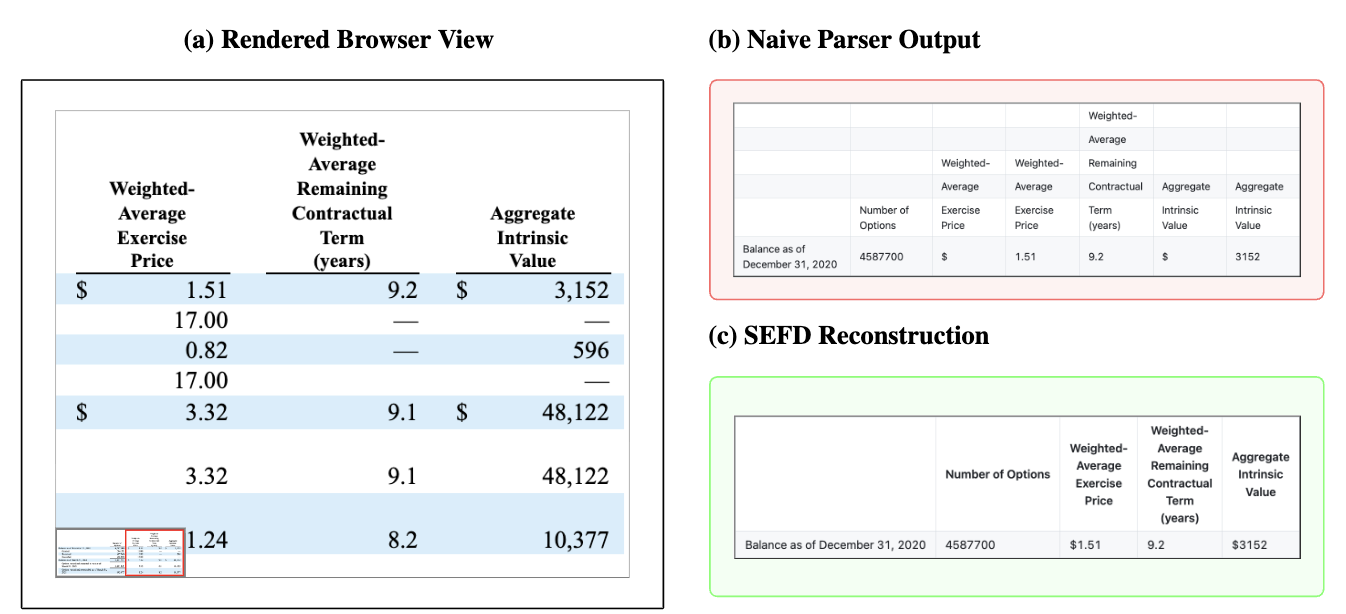}
    \caption{\textbf{Fragmented Headers.} Filing agents encode one visual header as rows. Browsers preserve grouping (a), while standard parsers fragment it (b) and SEFD reconstructs the logical header (c).}
    \label{fig:fragmented_header}
\end{figure}

\noindent \textbf{Header Reconstruction.} Although EDGAR permits inline tags like \texttt{<br>}, relying on them for multi-line headers can introduce rendering inconsistencies across legacy viewers. Filing agents therefore favor a row-by-row approach, exploding a single semantic header (see Figure~\ref{fig:fragmented_header}) into distinct table rows, each with its own cell-level attributes. This hard-codes vertical layout for browsers, but disconnects semantically connected elements and wastes tokens. SEFD reverse-engineers this structure using \texttt{border-*} and \texttt{margin-*} styling cues, then filters candidate headers by row cardinality and content density while excluding grids containing financial indicators such as currency symbols or voting data. When a fragmented header is confirmed, the rows are coalesced into a unified text block; when embedded sub-headers or body rows create ambiguity, the parser aborts reconstruction and falls back to deterministic row-by-row extraction.

\paragraph{Benefits of Structured MultiMarkdown.}
We also evaluate whether each HTML table representation preserves enough structure to reconstruct the original table. On 100 complex EDGAR HTML tables, GPT-5.4 (xhigh) \cite{openai:gpt54} is given only the parsed representation produced by SEFD (MMD), EdgarTools (Markdown), or a generic \texttt{to\_markdown} baseline, and is asked to reconstruct the table as HTML. SEFD MMD achieves 94.5\% adjusted recall, compared with 75.7\% for EdgarTools and 70.4\% for \texttt{to\_markdown}. This indicates that SEFD's reconstruction choices, combined with MMD span notation, preserve enough table structure for the reconstruction model to infer the original source layout while remaining token-efficient (Table~\ref{tab:html-reconstruction-benchmark}).

\subsection{XML}

\noindent As shown in Figure~\ref{fig:dataset-source-format-over-time}, XML is absent from the pre-2003 sample but appears sharply after the SEC's mid-2003 XML adoption for Section 16 ownership reports, reaching 39.4\% of source-format tokens by 2019-2025. This reflects the SEC's broader shift toward schema-validated, machine-readable disclosures, beginning with Forms 3/4/5 and expanding over time to additional high-cardinality submission types. Unlike HTML, where meaning is often implicit in layout, XML encodes disclosures as explicit fields governed by XSD schemas. For example, \texttt{<issuer>}/\texttt{<issuerTradingSymbol>} in Form 4 always refers to the issuer's stock ticker, independent of filing-agent formatting.

However, XML's field structure is not itself a pretraining document, as original document structure must be recreated from schema elements, optional branches, and repeated records. This is non-trivial for extensive filings such as Form N-PORT, where sections vary across files and the technical specification contains 89 explicit conditional or mutually exclusive branching rules.\footnote{Calculated as the unique conditional-mandatory (\texttt{m\#}) and mutually-exclusive choice (\texttt{c\#}) constraint IDs enumerated in the N-PORT XML Technical Specification's 'Applicability of Schema Elements' section (March 2025) \cite{sec:nport-techspec-1.13}: \texttt{m\#1}--\texttt{m\#71} and \texttt{c\#1}--\texttt{c\#18} (89 total). We exclude optional-conditional rules (\texttt{o\#1}--\texttt{o\#7}), which only apply when optional subtrees are present.} Given these schema combinatorics, SEFD prioritizes semantically significant and frequently occurring structures while acknowledging that rare branches or highly atypical conditional paths may remain unmapped. The 33 supported XML schemas fall into four broad archetypes: transactional and ownership disclosures (Forms 3, 4, 5, 13F, 13D/G, 144), fund and portfolio reporting (N-PORT, N-CEN, N-MFP, N-PX), primary offerings and issuer notices (Form D, 1-A/K/Z, C, ABS-EE), and entity registration and compliance filings (Form MA, TA-1/2, ATS-N, X-17A-5).

\subsection{SGML}

\noindent In SEFD, SGML functions in three capacities. First, every submission is wrapped in a \texttt{<SEC-HEADER>} block containing indexing metadata such as CIK and filing type, acting as a standardized wrapper from EDGAR's original ASCII/SGML era.\footnote{SEFD also parses legacy \texttt{<IMS-HEADER>} wrappers, a pre-2000s variant found in early electronic filings such as Form ADV or MSDW before the modern \texttt{<SEC-HEADER>} standard was universal.} Second, tags such as \texttt{<DOCUMENT>} and \texttt{<TYPE>} segment a submission into constituent parts, allowing the pipeline to separate the primary disclosure from auxiliary attachments such as XML information tables or uu-encoded PDFs. Third, SGML serves as a primary data container for investment company filings, including Form 497 and Form 24F-2NT. In these documents, \texttt{<SERIES-AND-CLASSES-CONTRACTS-DATA>} provides a hierarchy that groups Share Classes and Ticker Symbols under their parent Series ID, reflecting the SEC's 2006 move to require investment company Series and Class identifiers in EDGAR headers. Unlike administrative headers, these blocks contain valuable data that often does not appear in the narrative body.

\subsection{PDFs}

\noindent The inclusion of PDF files in EDGAR filings was formalized in May 1999 under Regulation S-T Rule 104. Originally designated as 'unofficial' copies, PDFs supplemented the official ASCII or HTML text by preserving visual material that early web standards could not render well, including glossaries, charts, and signatures. Their role has since expanded: as of January 2023, the SEC mandates PDF as the official format for ``glossy'' annual reports (Form ARS), making PDFs the primary record for some highly visual disclosures.

\paragraph{Extraction.}
PDFs enter SEFD through two paths. Some are explicit attachments, including material agreements, investor presentations, correspondence, signatures, and Form ID notarized authenticating documents. Others appear indirectly as ``PDF artifacts'': HTML generated by PDF-to-HTML converters, often as absolute-positioned \texttt{<div>} elements that bypass HTML~3.2 layout restrictions. SEFD processes PDF attachments with Mistral OCR 3 \cite{mistral:ocr3} in 10-page batches, skips nearly blank pages using a pixel-variance filter, requests HTML table output, and converts those HTML fragments into compact MultiMarkdown. This keeps PDF-derived tables compatible with the same downstream representation used for reconstructed HTML tables.

This design choice also motivated EDGAR-OCR. Although Qwen3.6-35B-A3B outperformed Mistral OCR 3 in accuracy (75.78\% vs. 75.33\%), Mistral was the Pareto-optimal production choice, reaching nearly the same score at roughly one-eleventh the median total latency (2.29s vs. 24.58s). PDF-sourced tokens constitute less than 2\% of the corpus, but they add non-redundant visual and semantic diversity across roughly 6.8M pages that would otherwise be lost in a text-only extraction pipeline.

\section{Dataset Analysis}

We analyze a 3.0B-token archive-wide SEFD sample to characterize the composition of the corpus by filing type, filing year, source format, and document length. Extrapolating from the sample yields 18.5M filings, 19.8TB of raw EDGAR source files, 1.9TB of parsed SEFD MultiMarkdown, and 548.9B parsed tokens.

\begin{figure}[H]
\centering

\begin{minipage}{0.88\linewidth}
\centering
\begin{subfigure}[t]{0.47\linewidth}
    \centering
    \includegraphics[width=\linewidth, trim=0 0 0 0, clip]{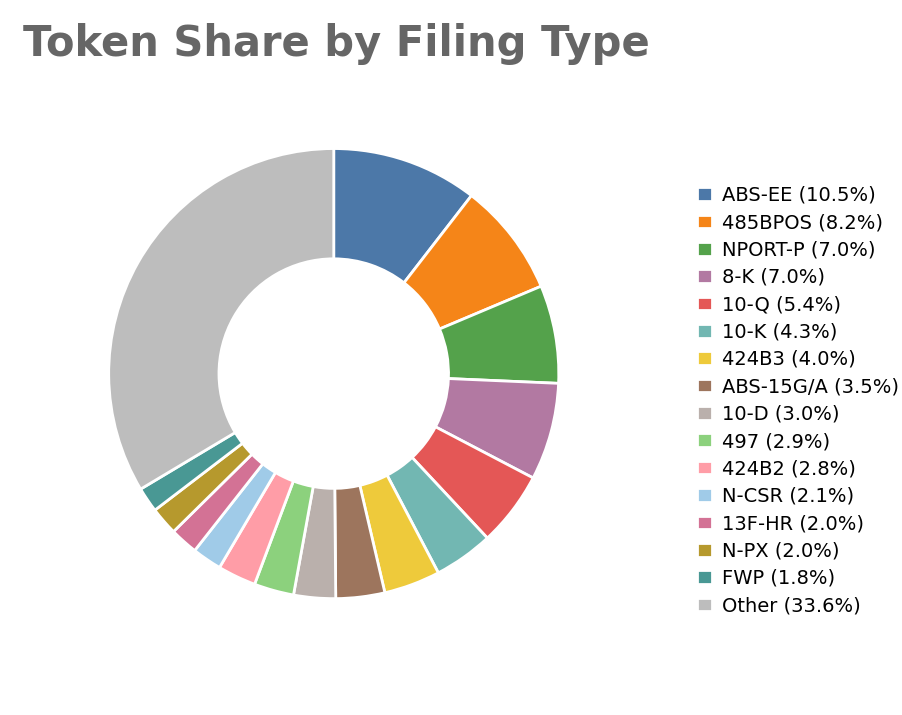}
    \caption{Filing-type token share.}
    \label{fig:dataset-form-type-token-share}
\end{subfigure}%
\hspace{0.06\linewidth}%
\begin{subfigure}[t]{0.47\linewidth}
    \centering
    \includegraphics[width=\linewidth, trim=7 6 8 8, clip]{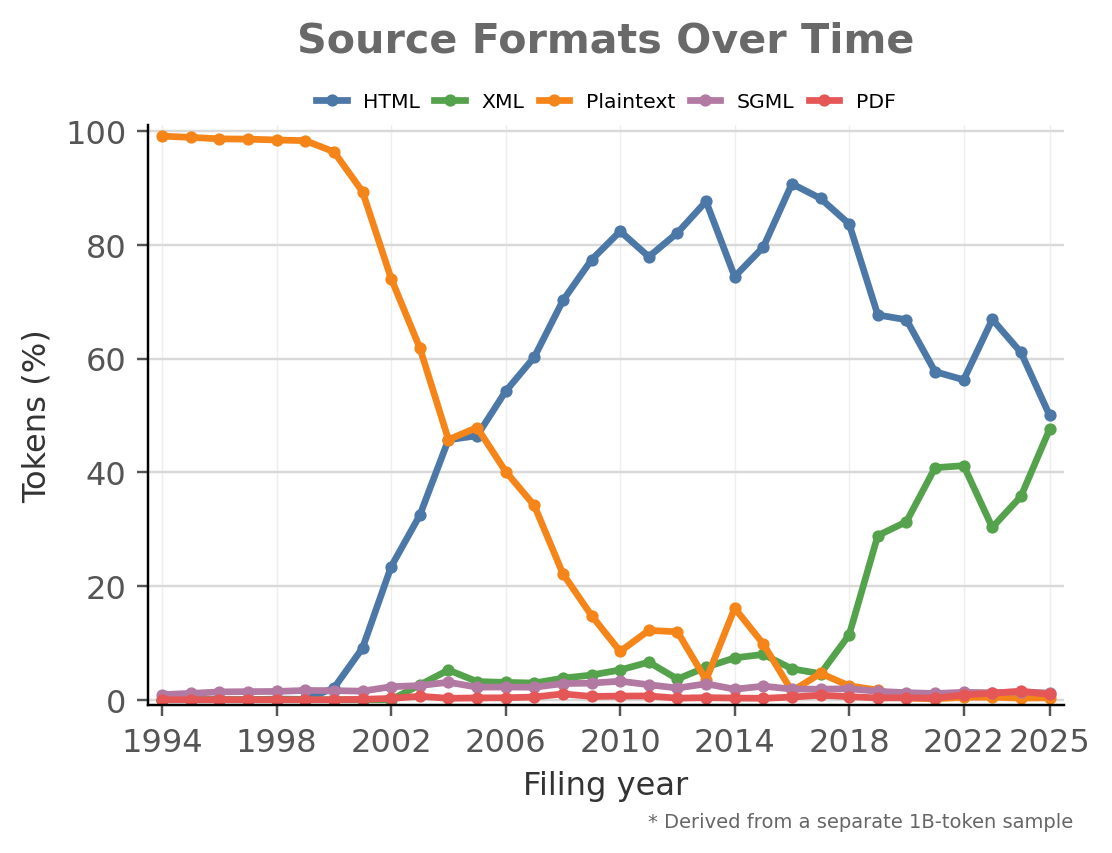}
    \caption{Source formats over time.}
    \label{fig:dataset-source-format-over-time}
\end{subfigure}
\end{minipage}

\vspace{0.7em}

\begin{minipage}{0.88\linewidth}
\centering
\begin{subfigure}[t]{0.47\linewidth}
    \centering
    \includegraphics[width=\linewidth, trim=8 8 8 8, clip]{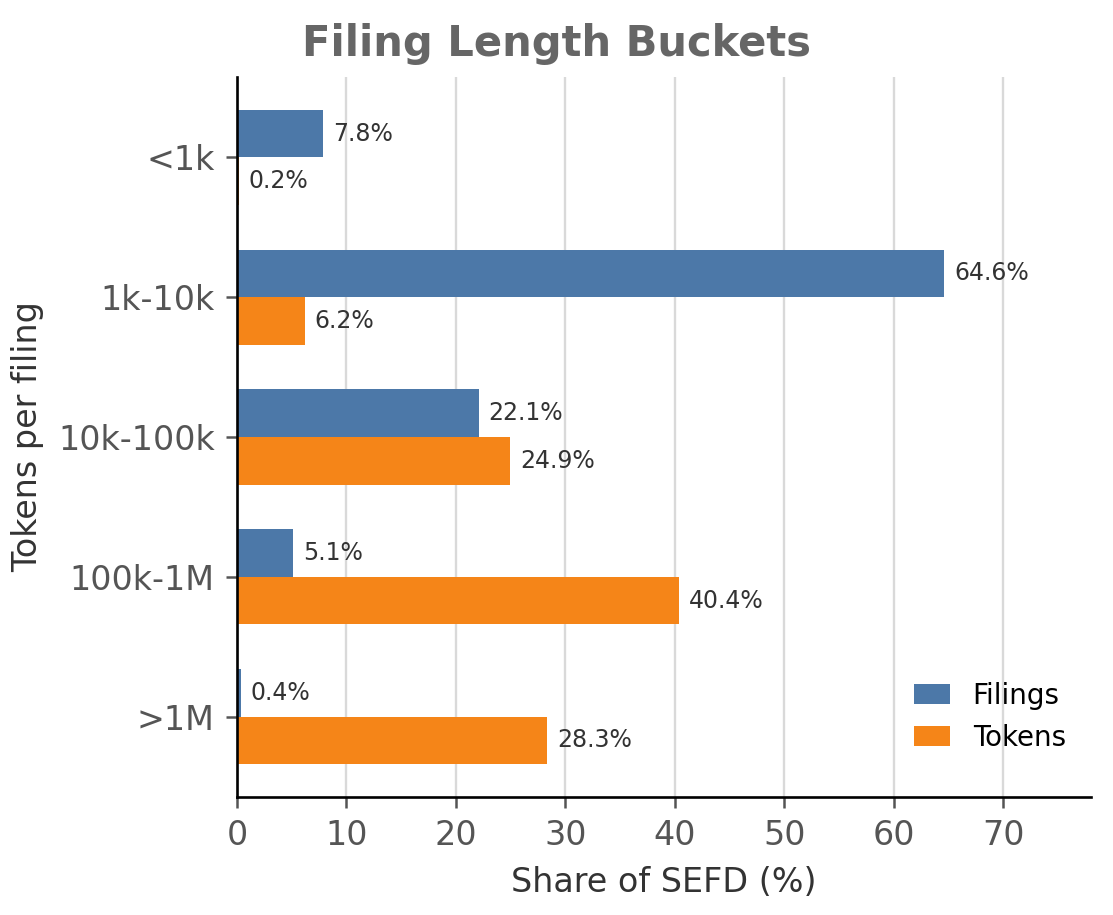}
    \caption{Filing-length token concentration.}
    \label{fig:dataset-token-length-buckets}
\end{subfigure}%
\hspace{0.06\linewidth}%
\begin{subfigure}[t]{0.47\linewidth}
    \centering
    \includegraphics[width=\linewidth, trim=8 8 8 8, clip]{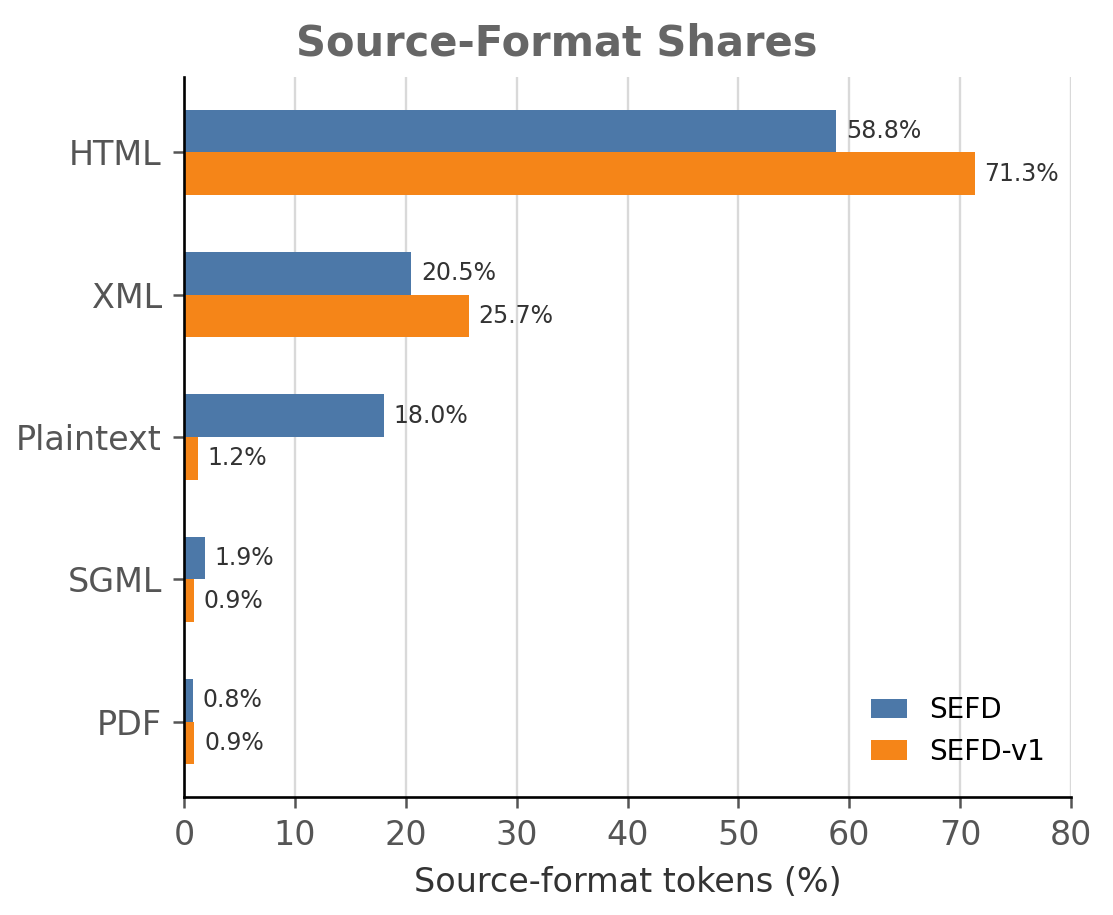}
    \caption{SEFD vs. SEFD-v1 source-format shares.}
    \label{fig:dataset-source-format-comparison}
\end{subfigure}
\end{minipage}

\caption{Dataset composition by filing type, source format, length, and SEFD-v1 source-format mix.}
\label{fig:dataset-analysis}
\end{figure}

The corpus is not dominated solely by canonical periodic reports. The five largest filing types account for 38.0\% of sampled tokens, led by ABS-EE, 485BPOS, NPORT-P, 8-K, and 10-Q filings (Figure~\ref{fig:dataset-form-type-token-share}). Standard 10-K and 10-Q reports account for 9.6\% of tokens, while Form 4 filings account for roughly one quarter of sampled filings but only 1.7\% of tokens. This highlights a recurring theme of EDGAR: filing frequency and training mass follow very different distributions.

The source-format mix changes sharply over time (Figure~\ref{fig:dataset-source-format-over-time}). Early EDGAR filings are almost entirely plaintext: in 1994, plaintext contributes 98.7\% of source-format tokens. HTML becomes dominant in the 2000s, rising from 24.1\% of source-format tokens in 2002 to 83.5\% by 2010. In recent years, XML/XBRL becomes a major component, as XML rises from 30.1\% in 2018 to 37.8\% in 2022 and 44.0\% in 2025. These transitions require SEFD to combine parsing strategies across multiple generations of SEC filing formats.

\begin{figure}[H]
    \centering
    \includegraphics[width=1\linewidth]{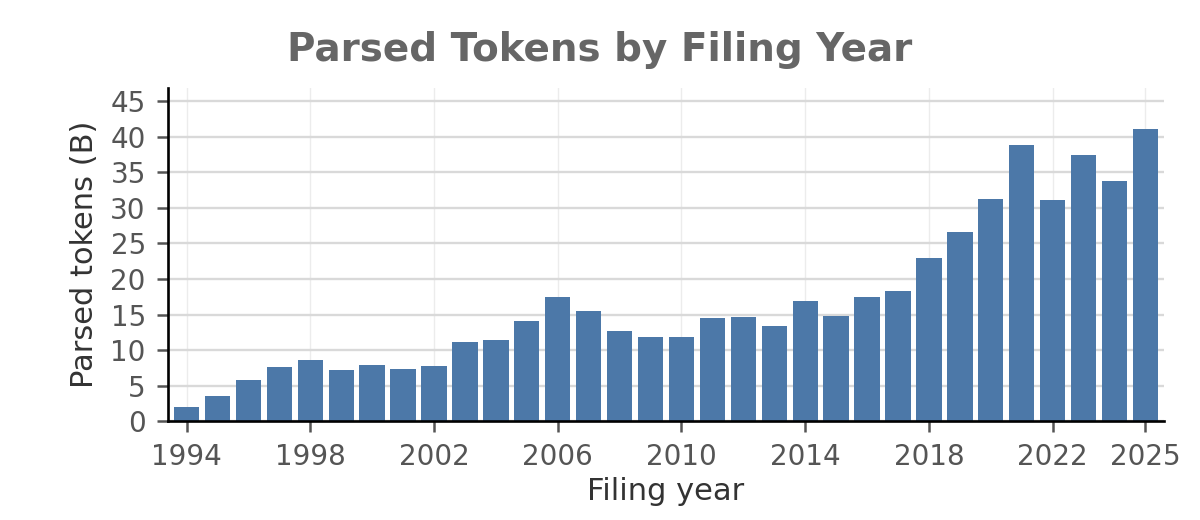}
    \caption{Annual parsed SEFD token volume by filing year, extrapolated from the 3.0B-token sample.}
    \label{fig:annual-token-volume}
\end{figure}

Filing lengths are also highly skewed (Figure~\ref{fig:dataset-token-length-buckets}). Filings with 1k--10k tokens account for 64.6\% of filings but only 6.2\% of tokens. In contrast, filings above 100k tokens account for just 5.5\% of filings but 68.7\% of tokens. Annual scale also changes unevenly (Appendix Figure~\ref{fig:annual-token-volume}): estimated token volume rises from 7.7B tokens in 2002 to 11.2B in 2003, and from 26.6B in 2019 to 31.3B in 2020, peaking at 41.1B in 2025.

Finally, SEFD-v1 differs from the archive-wide sample because it is concentrated in recent filings (Figure~\ref{fig:dataset-source-format-comparison}): HTML and XML rise to 71.3\% and 25.7\% of tokens, while plaintext falls to 1.2\%.

\section{EDGAR-OCR}

A nontrivial fraction of EDGAR filings contain tables that enter the archive as PDFs or image-like exhibits rather than clean HTML. We therefore evaluate OCR systems on the specific task needed by SEFD -- converting visually rendered SEC tables into faithful HTML-style table transcriptions, preserving cell placement, row and column structure, and inline formatting where the source table contains it. Example complex OCR inputs are shown in Appendix~\ref{app:edgarocr-examples}.

\paragraph{Benchmark Construction.}
We constructed an EDGAR-OCR benchmark from 300 hand-selected tables from raw SEC filings. Rather than evaluating models on the original filing tables, each selected table is converted into a synthetic benchmark instance to reduce train-test contamination and prevent models from recovering answers through memorization or external lookup. The pipeline parses the source table into MMD, deidentifies entity names and values, and uses GPT-5.4 (xhigh) to extract row-level variables and relationships from the table. These variables are then used to replace numeric entries while preserving local arithmetic. The resulting synthetic table is rendered back into filing-style HTML, screenshotted, and used as the OCR input, while the target transcription is generated from the same synthetic table before rendering.

\paragraph{Alternate Correct Tables.}
Some SEC tables admit multiple valid transcriptions, especially when cell boundaries are not directly visible or nested multi-row headers support more than one reasonable structure. We avoid penalizing correct alternatives with multiple-answer truth sets; across 300 samples, 241 have at least one alternate valid table, with 782 alternate transcriptions total. Each model output is scored against the primary transcription and all alternates, keeping the highest score.

\paragraph{Scoring.}
Predicted tables are parsed into cell grids and compared against the ground-truth grid. For each nonempty ground-truth cell, a prediction receives 1.0 point when the cell content, inline formatting signature, and grid placement all match. It receives 0.5 points when the text is correct but the inline formatting differs, and 0.25 points when the correct content appears in the predicted table but in the wrong cell. We report this adjusted-recall score as the main benchmark metric, along with inline formatting preservation and median latency.

\begin{figure}[H]
    \centering
    \includegraphics[width=0.68\linewidth]{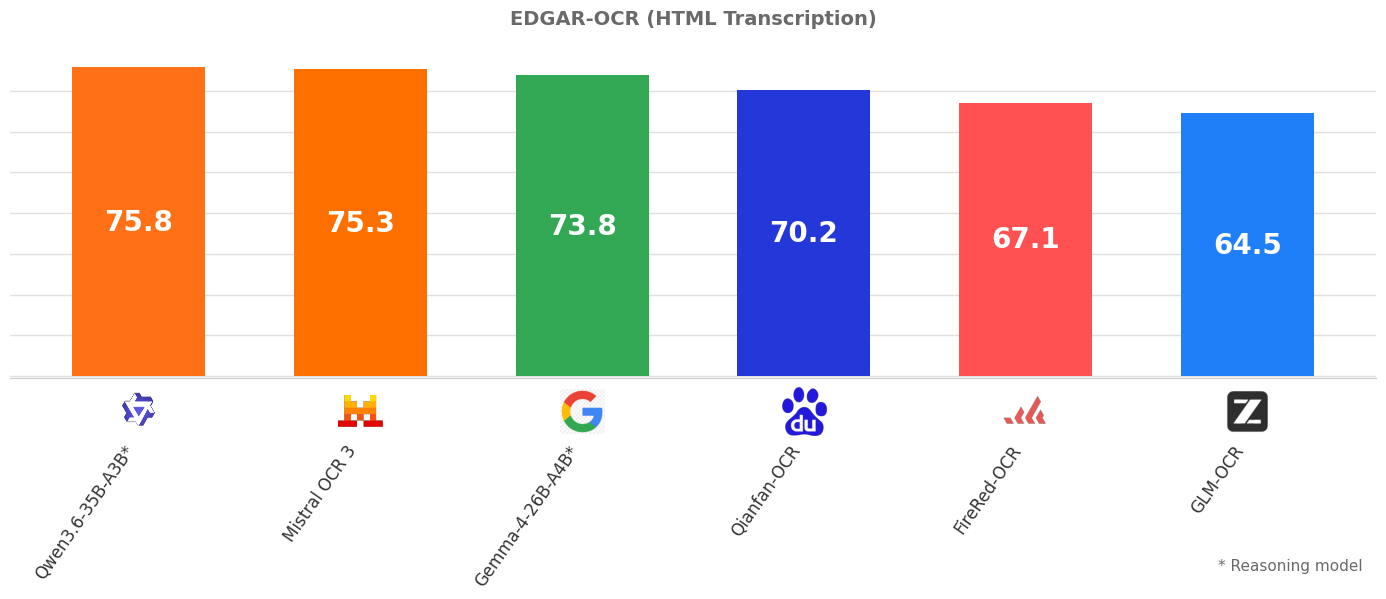}
    \caption{EDGAR-OCR adjusted recall on 300 hand-selected SEC tables.}
    \label{fig:edgarocr-results}
\end{figure}
\vspace{-1.75em}
\paragraph{Results.}
Qwen3.6-35B-A3B \cite{qwen:qwen36} obtains the highest adjusted recall, while Mistral OCR 3 \cite{mistral:ocr3} is nearly tied and substantially faster, motivating our use of Mistral OCR 3 in the high-throughput SEFD pipeline. Formatting remains a meaningful limitation across systems, as models that do not emit recoverable inline markup can score well on text and placement while still failing to preserve bold, italic, underline, superscript, or subscript structure. Additional latency and inline-formatting results are reported in Appendix~\ref{app:edgarocr-addresults}.

\section{EDGAR-Forecast}

EDGAR-Forecast evaluates whether models can use historical SEC filings to forecast numeric values that appear in a later hidden filing. Each benchmark instance is built around a company with a target 2026 10-Q filing. The model is given access to the company's prior filing history, but the target filing itself is hidden. The task is to predict five numeric values from the hidden filing, such as revenues, expenses, volumes, percentages, per-share quantities, or recurring financial and operating metrics.

\paragraph{Benchmark Construction.}
We construct EDGAR-Forecast using GPT-5.5 (xhigh) \cite{openai:gpt55} to generate candidate forecast targets. For each company, GPT-5.5 is run in the same local harness used for evaluation, except that the target filing is visible during construction. The model is instructed to select targets requiring holistic use of the prior five-year filing history, where valid targets require identifying relevant historical numbers, understanding the surrounding business prose, and using recurring trends, seasonality, segment behavior, guidance, or operating drivers to form a defensible forecast. Each candidate must include the target line, ground-truth value, consulted historical accessions, and evidence explaining why the value is forecastable from prior filings. Candidates are deterministically rejected if they repeat an already selected target, leak the exact answer from the visible five-year filing history, or appear to be direct copy-forward values from prior filings. We then manually review the resulting benchmark instances and exclude low-quality samples. The final benchmark contains 50 company-level instances and 250 targets; examples appear in Appendix~\ref{app:edgar-forecast-examples}.

\paragraph{Evaluation Harness.}
All models are evaluated through the same open-source Codex harness, which provides a standardized local file-system interface to the filing sandbox. The harness exposes only prior same-issuer filings, a filing index, and basic local file/search operations; the target filing, answers, and scorer remain outside the sandbox. No model is allowed web access or external data. Each model must inspect the same visible filing history and return predictions in the same JSON schema. All models are evaluated using their maximum available reasoning setting.

\paragraph{Scoring.}
Each benchmark instance contains five forecast targets and is scored out of five points. Each target receives 1 point for a prediction within the full-credit tolerance, 0.5 points for a prediction within the relaxed tolerance, and 0 points for a miss, missing answer, or non-numeric prediction. The reported score is the percentage of total available target credit; tolerances are listed in Appendix~\ref{tab:edgarscoring}.

\begin{figure}[H]
    \centering
    \includegraphics[width=0.62\linewidth]{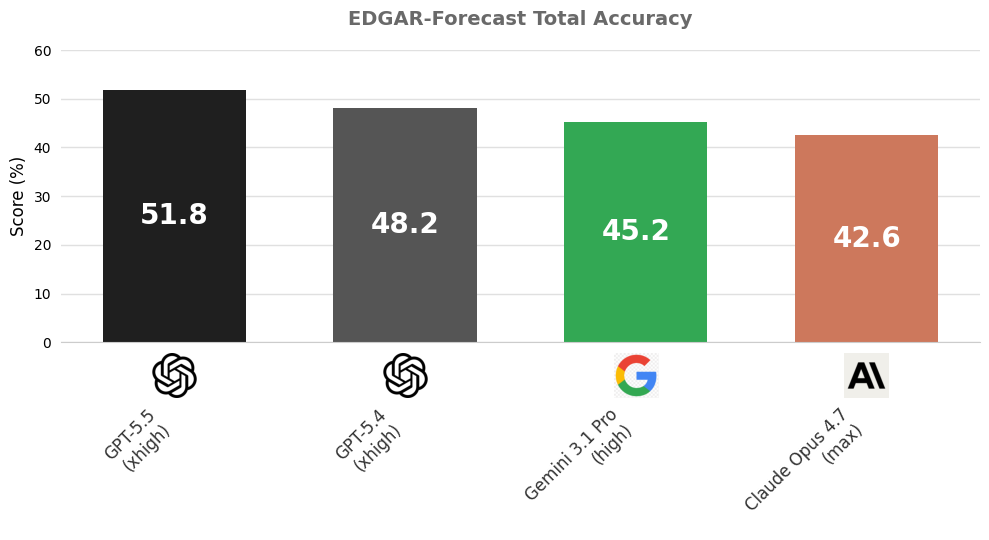}
    \vspace{-0.2em}
    \caption{EDGAR-Forecast accuracy across 50 company-level instances and 250 numeric targets.}
    \label{fig:edgar-forecast-accuracy}
\end{figure}

\paragraph{Results.}
GPT-5.5 \cite{openai:gpt55} achieves the highest overall score at 51.8\%, followed by GPT-5.4 \cite{openai:gpt54} at 48.2\%, Gemini 3.1 Pro \cite{google:gemini31pro} at 45.2\%, and Claude Opus 4.7 \cite{anthropic:opus47} at 42.6\% (Figure~\ref{fig:edgar-forecast-accuracy}). The task remains difficult even for frontier models because it combines planning, retrieval, and contextual reasoning -- models must answer five targets per filing, find relevant historical values, interpret surrounding prose and table context, and synthesize those signals into a filing-grounded forecast.

Additional analyses in Appendix~\ref{tab:edgaraddresultsscorestokens} show that scores are positively correlated with visible company filing history for all four models, most strongly for GPT-5.5 ($r=+0.38$), followed by Claude Opus 4.7 ($r=+0.22$), GPT-5.4 ($r=+0.20$), and Gemini 3.1 Pro ($r=+0.15$). The benchmark is token-intensive, with total evaluation usage ranging from 131.6M tokens for GPT-5.4 to 188.8M for Claude Opus 4.7 (Appendix~\ref{tab:edgaraddresultstokenusage}), mostly from cached inputs over long filing histories. GPT-5.5 provides the best accuracy-token tradeoff, achieving the highest score while using fewer total tokens than Gemini 3.1 Pro or Claude Opus 4.7.

\section{Conclusion}

We introduce SEFD, a layout-faithful reconstruction of the SEC EDGAR archive into token-efficient MultiMarkdown. SEFD converts heterogeneous plaintext, HTML, XML, SGML, and PDF filings into a standardized representation that preserves financial tables, indentation, document hierarchy, and visual structure often lost in text-only extraction. The resulting corpus provides an open source of long-context financial documents with low overlap against Common Crawl-derived corpora and is comparable to roughly 1.3B pages of reconstructed filing content.

Our analysis shows that SEFD's training signal is highly uneven across filing type, source format, and length. High-frequency forms are not necessarily high-token forms, and a small tail of long filings contributes most corpus mass. This heterogeneity makes the archive valuable for language modeling, but also difficult to reconstruct. SEFD's format-specific parsing pipeline addresses this challenge at scale, preserving both machine-readable disclosures and visually expressed financial structure.

Finally, EDGAR-OCR and EDGAR-Forecast show that high-fidelity filing reconstruction supports evaluation beyond retrieval-oriented search. EDGAR-OCR tests transcription of complex financial tables from visual inputs, while EDGAR-Forecast tests whether frontier models can use long historical filing context for filing-grounded numerical forecasts. The corpus and benchmarks showcase SEFD as a practical resource for financial language modeling, long-context pretraining, financial reasoning, document understanding, and evaluation.

\section*{Acknowledgments}
The SEFD-v1 production parse was run in part on Stanford's Marlowe GPU-based computational instrument \cite{kapfer2025marlowe}.

\section{Appendix}

\subsection{Markup Language Comparison}

\FloatBarrier

\subsection{SGML Header Example}

\begin{lstlisting}[label={lst:sgml_header},caption={SGML header example for AAPL},captionpos=b]
<SEC-DOCUMENT>0001193125-15-023732.txt : 20150128
<SEC-HEADER>0001193125-15-023732.hdr.sgml : 20150128
<ACCEPTANCE-DATETIME>20150128164846
ACCESSION NUMBER:		0001193125-15-023732
CONFORMED SUBMISSION TYPE:	8-K
PUBLIC DOCUMENT COUNT:		11
CONFORMED PERIOD OF REPORT:	20150128
ITEM INFORMATION:		Other Events
ITEM INFORMATION:		Financial Statements and Exhibits
FILED AS OF DATE:		20150128
DATE AS OF CHANGE:		20150128

FILER:

	COMPANY DATA:	
		COMPANY CONFORMED NAME:			APPLE INC
		CENTRAL INDEX KEY:			0000320193
		STANDARD INDUSTRIAL CLASSIFICATION:	ELECTRONIC COMPUTERS [3571]
		IRS NUMBER:				942404110
		STATE OF INCORPORATION:			CA
		FISCAL YEAR END:			0927

	FILING VALUES:
		FORM TYPE:		8-K
		SEC ACT:		1934 Act
		SEC FILE NUMBER:	001-36743
		FILM NUMBER:		15555430

	BUSINESS ADDRESS:	
		STREET 1:		ONE INFINITE LOOP
		CITY:			CUPERTINO
		STATE:			CA
		ZIP:			95014
		BUSINESS PHONE:		(408) 996-1010

	MAIL ADDRESS:	
		STREET 1:		ONE INFINITE LOOP
		CITY:			CUPERTINO
		STATE:			CA
		ZIP:			95014

	FORMER COMPANY:	
		FORMER CONFORMED NAME:	APPLE COMPUTER INC
		DATE OF NAME CHANGE:	19970808
</SEC-HEADER>
\end{lstlisting}

\subsection{HTML}

\subsubsection{Additional HTML Parsing Methods}
\label{app:html-layout-signals}

\paragraph{Visual Indentation Normalization.}
Indentation in SEC filings is a strong indicator of hierarchical relationship (e.g., 'Cash' indented under 'Current Assets'). Lacking a standardized semantic tag for hierarchy, filers achieve visual offset through non-breaking spaces (\texttt{\&nbsp;}), fixed-width empty columns, or \texttt{<p>} tags with varying CSS attributes (\texttt{margin-left}, \texttt{padding-left}) in inconsistent units (\texttt{px}, \texttt{pt}, \texttt{em}). SEFD normalizes these inputs into a unified 'indentation level' by calculating an indent value for every text block, converting CSS units to points (e.g., 1\texttt{em} = 12\texttt{pt}), and binning the result into discrete levels represented as non-breaking spaces. This preserves the hierarchy of financial statements independent of the filing software's implementation.

\paragraph{Font Styling Preservation.}
Financial disclosures rely on font styling to denote semantic weight (\textbf{bold} (\texttt{<b>}), \textit{italic} (\texttt{<i>}), \underline{underline} (\texttt{<u>})). Naive parsers strip HTML tags during text processing, flattening formatted elements into plain text. SEFD instead uses a 'Placeholder Injection' pass before DOM parsing, wherein unique semantic tokens (e.g., \texttt{\#\#BOLD\_START\_ID\#\#}, \texttt{\#\#BOLD\_END\_ID\#\#}) are injected as raw text around target tags. The same technique applies to superscripts, subscripts, rowspans, and colspans. Because these placeholders are treated as content rather than markup, they survive DOM traversal; because each token pair has a unique ID, the parser preserves the original document's styling deterministically.

\subsubsection{HTML Table Reconstruction Benchmark Additional Results}

\begin{table}[H]
\centering
\small
\begin{tabular}{lccc}
\toprule
Representation & Adjusted recall & Weighted recall & Exact shape \\
\midrule
SEFD MMD & 94.5\% & 93.2\% & 89.0\% \\
EdgarTools & 75.7\% & 72.0\% & 79.0\% \\
\texttt{to\_markdown} & 70.4\% & 64.7\% & 71.0\% \\
\bottomrule
\end{tabular}
\vspace{1em}
\caption{HTML table reconstruction accuracy on 100 synthetic SEC-style tables. Adjusted recall gives partial credit for recoverable content and placement, weighted recall measures nonempty-cell recovery, and exact shape measures whether the reconstructed table grid matches the ground-truth table shape.}
\label{tab:html-reconstruction-benchmark}
\end{table}

\subsubsection{Annual SEFD Token Volume}
\label{app:annual-token-volume}

\vspace{-1em}

\vspace{-2em}

\subsection{EDGAR-OCR}

\subsubsection{EDGAR-OCR Examples}
\label{app:edgarocr-examples}

\begin{figure}[H]
    \centering
    \includegraphics[width=\linewidth]{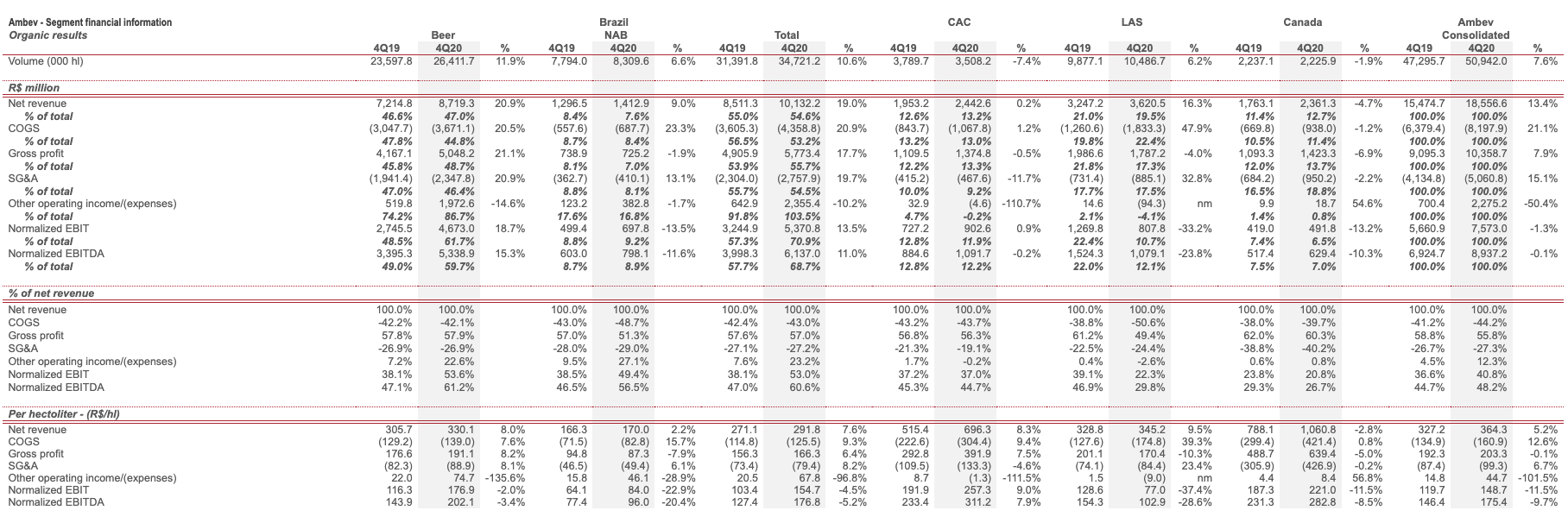}
    \caption{Example complex EDGAR table used in EDGAR-OCR. The table contains dense content, nested multi-row headers, and ambiguous merged cells whose span structure often requires financial-table reasoning rather than visual segmentation alone.}
\end{figure}

\vspace{-3em}

\begin{figure}[H]
    \centering
    \includegraphics[width=\linewidth]{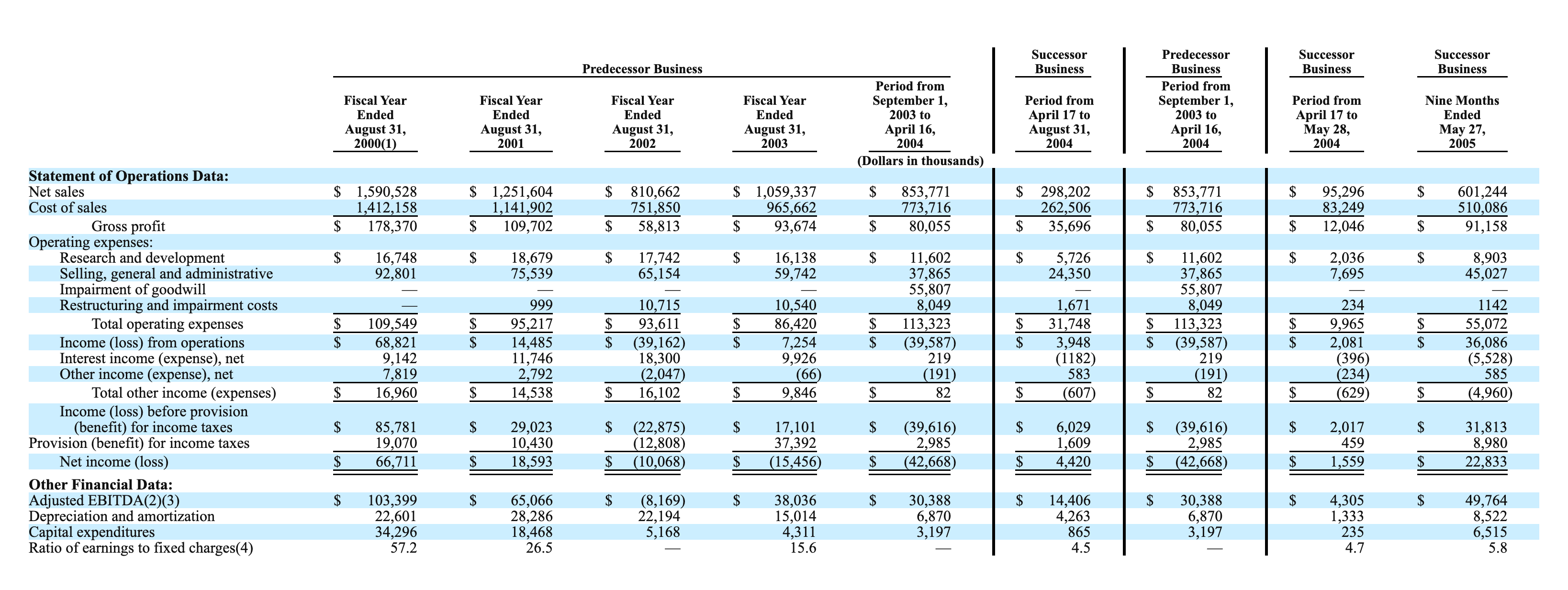}
    \caption{Example complex EDGAR table used in EDGAR-OCR. The table contains dense financial rows, grouped period headers, subtotal structure, and numeric formatting that require preserving both text content and table layout.}
\end{figure}

\subsubsection{EDGAR-OCR Additional Results}
\label{app:edgarocr-addresults}

\begin{figure}[H]
    \centering
    \begin{minipage}{0.48\linewidth}
        \centering
        \includegraphics[width=\linewidth]{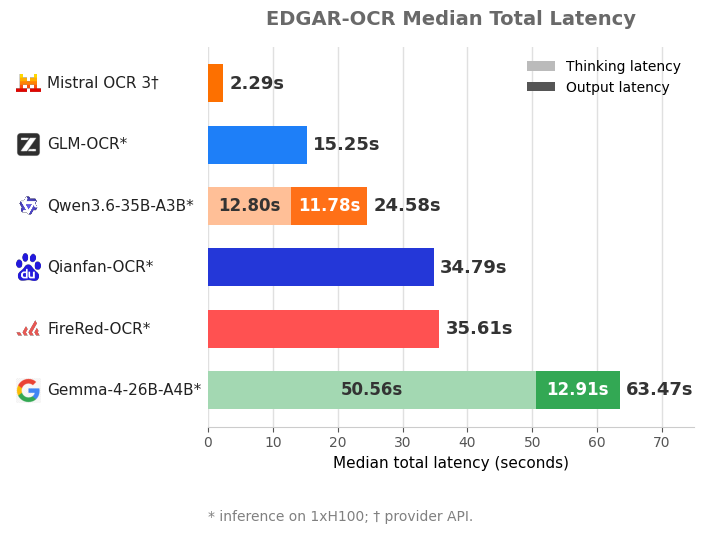}
        \caption*{(a) Median total latency.}
    \end{minipage}
    \hfill
    \begin{minipage}{0.48\linewidth}
        \centering
        \includegraphics[width=\linewidth]{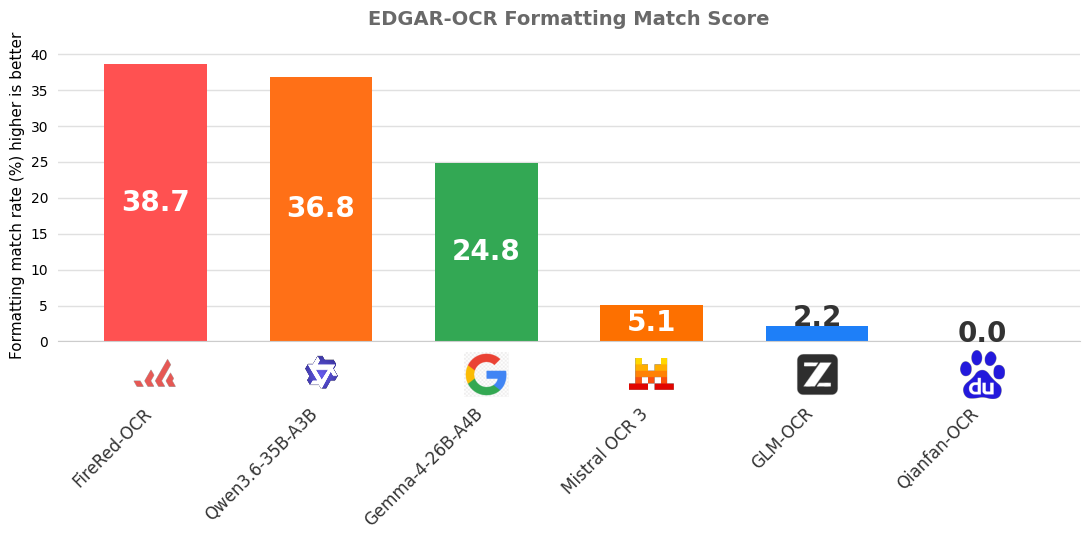}
        \caption*{(b) Inline formatting preservation.}
    \end{minipage}
    \caption{Additional EDGAR-OCR results: latency and inline formatting preservation.}
    \label{fig:edgarocr-latency-formatting}
\end{figure}

\subsection{EDGAR-Forecast}

\subsubsection{Example EDGAR-Forecast Questions}
\label{app:edgar-forecast-examples}

\begin{table}[H]
\centering
\small
\begin{tabular}{p{0.16\linewidth}p{0.17\linewidth}p{0.50\linewidth}p{0.10\linewidth}}
\toprule
Company & Accession & Forecast question & Type \\
\midrule
Tyson Foods & 0000100493-26-000007 & What is the value for Beef segment sales volume change, Q1 fiscal 2026 vs Q1 fiscal 2025? & Percent \\
Tyson Foods & 0000100493-26-000007 & What is the value for Cash outflow for additions to property, plant and equipment, Q1 fiscal 2026? & Amount \\
Digi International & 0000854775-26-000005 & What is the value for IoT Solutions operating income margin increase for the first quarter of fiscal 2026 versus the first quarter of fiscal 2025? & Bps \\
Powell Industries & 0000080420-26-000015 & What is the value for Electric utility market revenue, first quarter Fiscal 2026? & Amount \\
Powell Industries & 0000080420-26-000015 & What is the value for Revenue recognized during first quarter Fiscal 2026 from contract liabilities outstanding at September 30, 2025? & Amount \\
\bottomrule
\end{tabular}
\vspace{0.2em}
\caption{Representative EDGAR-Forecast questions from hidden 2026 10-Q filings, spanning segment volume, capital expenditures, margin expansion, end-market revenue, and contract-liability revenue recognition.}
\label{tab:edgar-forecast-example-questions}
\end{table}

\subsubsection{EDGAR-Forecast Scoring}
\label{tab:edgarscoring}

\begin{table}[H]
\centering
\small
\begin{tabular}{lcc}
\toprule
Target type & Full credit & Half credit \\
\midrule
Amounts, counts, per-share values & $\leq$5\% relative error & $\leq$10\% relative error \\
Percentages & $\leq$0.5 percentage points & $\leq$1.0 percentage point \\
Ratios & $\leq$0.05 absolute ratio points & $\leq$0.10 absolute ratio points \\
Basis points & $\leq$25 bps & $\leq$50 bps \\
\bottomrule
\end{tabular}
\vspace{0.2em}
\caption{EDGAR-Forecast scoring tolerances.}
\label{tab:edgar-forecast-scoring}
\end{table}

\subsubsection{EDGAR-Forecast Additional Results - Scores vs. Token Count}
\label{tab:edgaraddresultsscorestokens}

\begin{figure}[H]
    \centering
    \includegraphics[width=\linewidth]{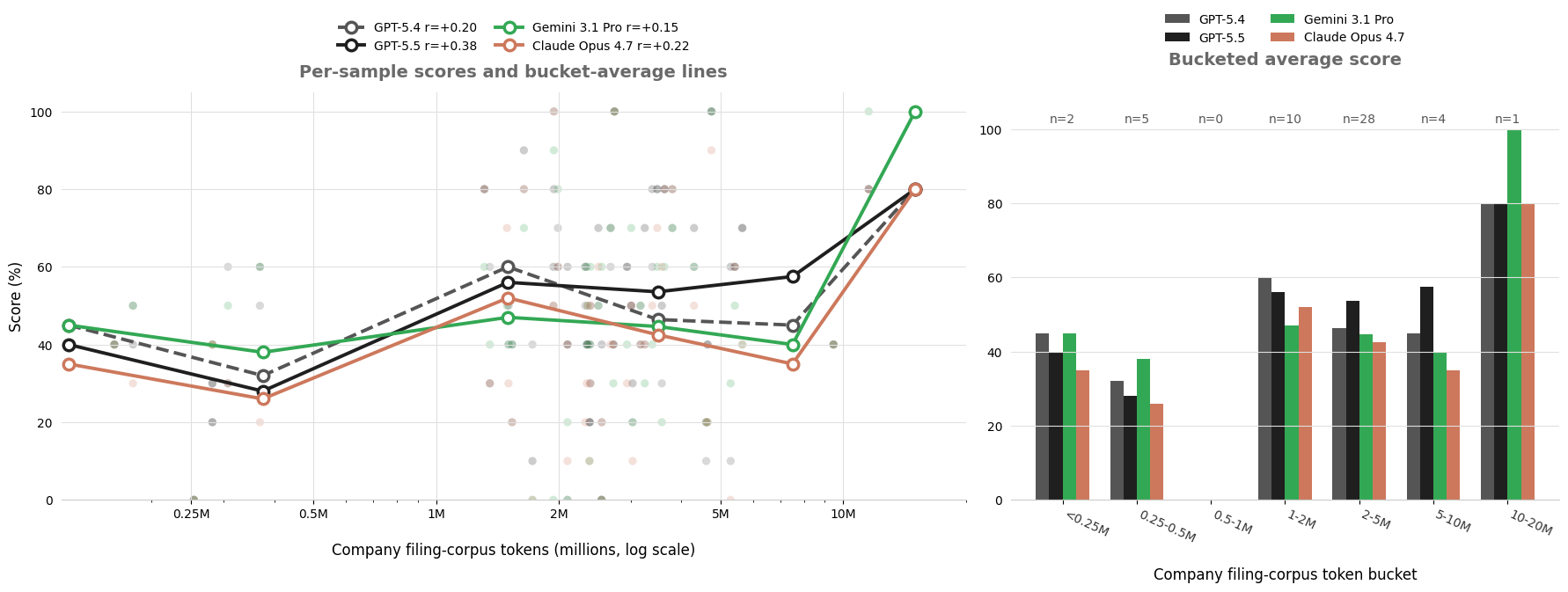}
    \caption{EDGAR-Forecast score as a function of visible company filing-history size. Dots show per-sample scores and lines show token-bucket averages.}
    \label{fig:edgar-forecast-score-vs-tokens}
\end{figure}

\subsubsection{EDGAR-Forecast Additional Results - Token Usage}
\label{tab:edgaraddresultstokenusage}

\begin{figure}[H]
    \centering
    \includegraphics[width=\linewidth]{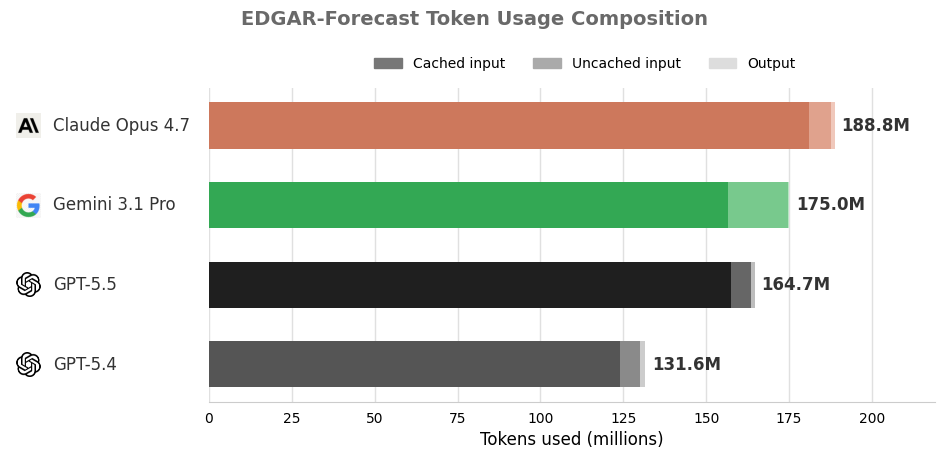}
    \caption{Total token usage for EDGAR-Forecast evaluation, decomposed into cached input, uncached input, and output tokens.}
    \label{fig:edgar-forecast-token-usage}
\end{figure}

\subsection{OCR Prompts}

\subsubsection{PDF OCR Extraction Prompt}

\begin{lstlisting}
"Extract the document content. Return prose and non-table text as Markdown."
"For every table, return real HTML table markup instead of Markdown tables."
"Use only plain table structure tags such as <table>, <tr>, <th>, and <td>,"
"with colspan and rowspan when needed to preserve merged cells."
"Do not wrap HTML tables in markdown code fences. Preserve visible table text,"
"row order, column order, punctuation, signs, and numeric formatting."
\end{lstlisting}

\subsubsection{HTML Reconstruction Prompt}

\begin{lstlisting}
"Reconstruct a single self-contained HTML <table> fragment from the parser markdown below.\n\n"
"Requirements:\n"
"- Preserve cell text, row/column layout, colspan, rowspan, and inline formatting when possible.\n"
"- Use only simple table HTML: <table>, <thead>, <tbody>, <tr>, <th>, <td>, <b>, <i>, <u>, <sup>, <sub>, <br>.\n"
"- Do not include CSS, classes, styles, scripts, surrounding prose, or explanation.\n"
"- Actively reconstruct colspan/rowspan when the markdown gives strong evidence of a merged header or merged body cell. If the alternative is losing an obvious grouped-header or merged-cell structure, reconstruct the required span.\n"
"- Use explicit merge cues when present.\n"
"{parser_specific_merge_hints}"
"- Do not infer a colspan or rowspan from repeated adjacent text alone.\n"
"- Ordinary empty cells are real empty cells.\n"
"- Use <th> for obvious header cells when the markdown makes them clear; otherwise use <td>.\n\n"
"Parser source: {parser_name}\n\n"
"Parser markdown table:\n"
"{parser_markdown}"
\end{lstlisting}

\subsubsection{Parser-Specific Merge Hints}

\begin{lstlisting}
SEFD MMD:
"- Treat adjacent empty cells implied by `||` as cells covered by a colspan from the left.\n"
"- Treat `^^` as a placeholder cell covered by a rowspan from above.\n"

EdgarTools:
"- EdgarTools parsing is a lossy view of the source table; reconstruct the most plausible source HTML table, not merely a flat HTML copy of the markdown.\n"
"- EdgarTools parsing may flatten merged headers or merged body cells; restore the most plausible colspan/rowspan structure when grouped headers or obvious span relationships are clearly implied.\n"
"- If a header hierarchy would become nonsensical without a merged cell, reconstruct the required span.\n"
"- If blank cells are best explained by a header or label continuing from above or from the left, reconstruct the corresponding rowspan or colspan.\n"

to_markdown:
"- Plain `to_markdown()` converts the literal HTML table through a generic dataframe-like markdown view, which is especially lossy about merged headers and blank span cells.\n"
"- Reconstruct the most plausible source HTML table, not merely a flat HTML copy of the markdown.\n"
"- If grouped headers or label continuations would otherwise become nonsensical, restore the required colspan or rowspan.\n"
"- If blank cells are best explained by a header or label continuing from above or from the left, reconstruct the corresponding rowspan or colspan.\n"
\end{lstlisting}

\section{Limitations}

SEFD does not currently parse embedded images, and PDF-derived content depends on OCR quality. Parser coverage may lag newly introduced or changed SEC filing schemas, especially for rare XML branches or atypical conditional paths. Extremely layout-heavy filings can still produce lower-fidelity reconstructions than standard HTML, XML, SGML, or plaintext filings.

\end{document}